\documentclass{article} 
\usepackage{iclr2017_conference,times}
\usepackage{hyperref}
\usepackage{url}
\usepackage{enumitem}

\usepackage{lipsum}
\usepackage{graphicx}
\usepackage{float}
\usepackage{amsmath,amsfonts,amssymb}
\usepackage{algorithm}
\usepackage[noend]{algpseudocode}

\usepackage[font=small]{caption}

\pdfoutput=1

\title{Learning recurrent representations for\\ hierarchical behavior modeling}

\author{Eyrun Eyjolfsdottir$^1$, Kristin Branson$^2$, Yisong Yue$^1$, \& Pietro Perona$^1$\\
$^1$California Institute of Technology, 
$^2$Janelia Research Campus HHMI\\
}

\begin{document}

\maketitle

\vspace{-3mm}
\begin{abstract}
We propose a framework for detecting action patterns from motion sequences and modeling the sensory-motor relationship of animals, using a generative recurrent neural network.
The network has a discriminative part (classifying actions) and a generative part (predicting motion), whose recurrent cells are laterally connected, allowing higher levels of the network to represent high level behavioral phenomena. We test our framework on two types of data, fruit fly behavior and online handwriting. Our results show that 1) taking advantage of unlabeled sequences, by predicting future motion, significantly improves action detection performance when training labels are scarce, 2) the network learns to represent high level phenomena such as writer identity and fly gender, without supervision, and 3) simulated motion trajectories, generated by treating motion prediction as input to the network, look realistic and may be used to qualitatively evaluate whether the model has learnt generative control rules.  
\end{abstract}

\vspace{-3mm}
\section{Introduction}
Behavioral scientists strive to decode the functional relationship between sensory input and motor output of the brain \citep{tinbergen1963aims, moore2002some}. In particular, ethologists study the natural behavior of animals while neuroscientists and psychologists study behavior in a controlled environment, manipulating neural activations and environmental stimuli. These studies require quantitative measurements of behavior to discover correlations or causal relationships between behaviors over time or between behavior and stimuli; automating this process allows for more objective and precise measurements, and significantly increased throughput \citep{anderson2014toward}. Many industries are also concerned with automatic measurement and prediction of behavior, for applications such as surveillance, assisted living, sports analytics, self driving vehicles, and robotic/virtual assistants.

Behavior is complex and may be perceived at different time-scales of resolution: position, trajectory, action, activity. While position and trajectory are geometrical notions, action and activity are semantic in nature. The analysis of behavior may therefore be divided into two steps: (a) detection and tracking, where the pose of the body over time is estimated, and (b) action/activity detection and classification, where motion is segmented into meaningful intervals, each one of which is associated with a goal or a purpose. 
Our work focuses on going from (a) to (b), that is to detect and classify actions from motion trajectories.  We use data for which tracking and pose estimation is relatively simple, which lets us focus on modeling the temporal dynamics of pose trajectories without worrying about errors stemming from low level feature extraction.

Supervised learning is a powerful tool for learning classifiers from examples of actions provided by an expert \citep{kabra2013jaaba,eyjolfsdottir2014detecting}. However, it has two drawbacks. First, it requires a lot of training examples which involves time consuming and painstaking annotation. Second, behavior measurement is limited to actions that a human can perceive and believes to be important. 
We propose a framework that takes advantage of both labeled and unlabeled sequences, by simultaneously predicting future motion and detecting actions, allowing the system to learn action classifiers from fewer expert labels and to discover unbiased behavior representations.

The framework models the sensory-motor relationship of an agent, predicting motion based on its sensory input and motion history. It can be used to simulate an agent by iteratively feeding motion predictions as input to the network and updating sensory inputs accordingly.  A model that can simulate realistic behavior has learnt to emulate the generative control laws underlying behavior, which could be a useful tool for behavior analysis \citep{simon1996sciences, braitenberg84}.

Our model is constructed with the goal that it will learn to represent and discover behaviors at different semantic scales, offering an unbiased way of measuring behavior with minimal human input. 
Recent work by \cite{berman2014mapping} and \cite{wiltschko2015mapping} shows promising results towards unsupervised behavior representation. Compared to their work our framework offers three advantages. Our model learns a \textbf{hierarchical} embedding of behavior, can be trained \textbf{semi-supervised} to learn specific behaviors of interest, and our sensory-motor representation enables the model to learn \textbf{interactive} behavior of an agent with other agents and with its environment.

Our experiments  focus mainly on the behavior of fruit flies, \textit{Drosophila Melanogaster}, a popular, relatively simple, model organism for the study of behavior \citep{siwicki2009fruitless}.  To explore the generality of our approach we also test our model on online handwriting data, an interesting human behavior that produces two dimensional trajectories.

To summarize our contributions: 
\vspace{-2mm}
\begin{enumerate}[leftmargin=5mm, label={\arabic*)}]
\setlength\itemsep{0.0em}
\item We propose a framework that simultaneously models the sensory-motor relationship of an agent and classifies its actions, and can be trained with partially labeled sequences.
\item We show that motion prediction is a good auxiliary task for action classification, especially when training labels are scarce.
\item We show that simulated motion trajectories resemble trajectories from the data domain and can be manipulated by activating discriminative cell units.
\item We show that the network learns to represent high level information, such as gender or identity, at higher levels of the network and low level information, such as velocity, at lower levels.
\item We test our framework on the spontaneous and sporadic behavior of fruit flies, and the intentional and structured behavior of handwriting. 
\end{enumerate}


\section{Background} 
\textbf{Hidden Markov models} (HMMs) have been extensively used for sequence classification. The motivating assumption for HMMs is that there exists a process that transitions with some probability between discrete states, each of which emits observations according to some distribution, and the objective is to learn these functions given a sequence of observations and states. This model is limited in that its transition functions are linear, state space is discrete, and emission distribution is generally assumed to be Gaussian, although generalizations of the model that fall under the category of dynamic Bayesian networks are more expressive \citep{murphy2002dynamic}.

\textbf{Recurrent neural networks} (RNNs) have recently been shown to be extremely successful in classifying time series data, especially with the popularization of long short term memory cells \citep{hochreiter1997long}, in applications such as speech recognition \citep{graves2013speech}. RNNs have also been used for generative sequence prediction of handwriting \citep{graves2013generating} as well as speech synthesis \citep{chung2015recurrent}.

\textbf{Imitation learning} involves learning to map a state to an action, from demonstrated sequences of actions. This is a supervised learning technique which, when implemented as an RNN, can be trained via backpropagation using action-error computed at every time step. The problem with this approach is that the domain of states that an agent is trained on consists only of states that the demonstrators encounter, and when an agent makes a mistake it finds itself in a situation never experienced during training. Reinforcement learning handles this by letting an agent explore the domain using an action policy, and updating the policy based on a goal-specific penalty or reward which may be obtained after taking several actions. This exploration can be extremely expensive, and therefore it is common to precede reinforcement learning with imitation learning to start the agent off with a reasonable policy. This strategy is used in \citep{mnih2015human} where an agent is trained to play Atari games, and in \citep{silver2016mastering} for mastering the game of GO.

\textbf{Autoencoders} \citep{rumenlhart1986learning} have been used in semi-supervised classification to pre-train a network on an auxiliary task, such as denoising, to prevent overfitting on a small number of labeled data \citep{baldi2012autoencoders}. Recent work in this area \citep{rasmus2015semi} proposes to train on the primary and auxiliary task concurrently and using lateral connections \citep{valpola2015neural} between encoding and decoding layers to allow higher layers of the network to focus on high level features.



Our framework takes inspiration from each of the works described here.

\section{Model} 
\label{model}
Our model is a recurrent neural network, with long short term memory, that simultaneously classifies actions and predicts future motion of \textit{agents} (insects, animals, and humans). 
Rather than actions being a function of the recurrent state, as is common practice, our model embeds actions in recurrent state units. This way the recurrent function encodes action transition probabilities and motion prediction is a direct function of actions, similar to an HMM. The network takes as input an agent's motion and sensory input at every time step, and outputs the agent's next move according to a policy, which is effectively learnt via imitation learning. Similar to autoencoders, our model has a discriminative path, used to embed high level information, and a generative path used to reconstruct the input domain, in our case filling in the future motion. Each discriminative recurrent cell is fully connected with its corresponding generative cell, allowing higher level states to represent higher level information, similar to the idea of Ladder networks \citep{valpola2015neural}. 

\begin{figure}[h]
  \centering
\vspace{1mm}
  \hspace{-2mm}
  \includegraphics[width=\textwidth]{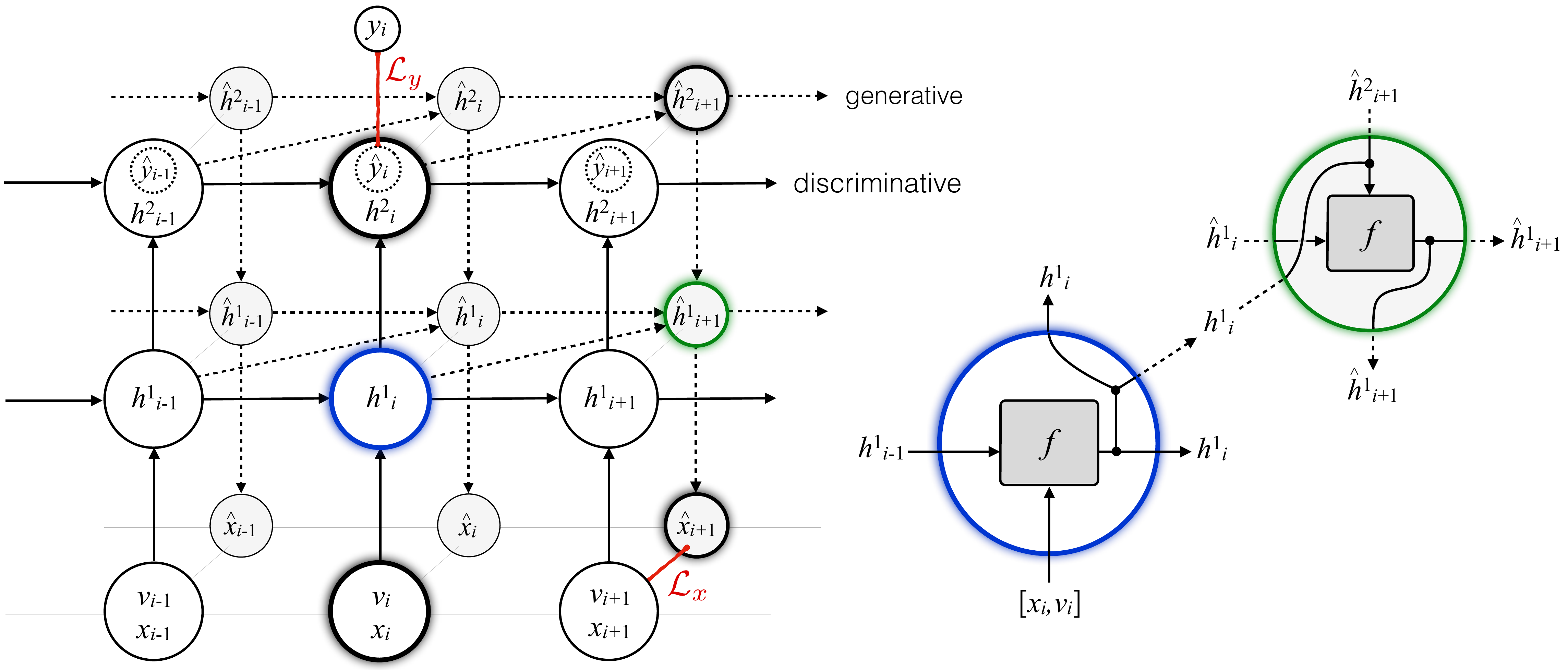}
  \hspace{2mm}
  \caption{\textbf{Left:} A 3D depiction of our network unrolled for 3 timesteps. The highlighted cells show the path from an input through a classification cell to a motion prediction output. During training, motion prediction loss $\mathcal{L}_x$ is computed at every timestep, and classification loss $\mathcal{L}_y$ is computed only at frames for which labels are provided. The diagonal connections between discriminative and generative cells enable higher levels of the network to represent high level information. Vector $v$ represents agent's sensory input, $x$ its motion, $h$ its internal state, and $y$ labeled actions. \textbf{Right:} A zoom in on the blue and green cells showing the recurrent state (horizontal arrows) and inputs to the recurrent cell function $f$. Merging of arrows represents vector concatenation, and branching vector duplication.}
    \label{mainfigure}
\end{figure}

\subsection{Architecture} 
The model can be thought of as two parallel recurrent networks: The \textit{discriminative} network takes as input an agent's motion, $x$, and environmental sensory input, $v$, and propagates them up through its hidden states which encode high level information, including action labels, $y$. The \textit{generative} network decodes the states of the discriminative network, propagating information down to predict the agents locomotion at the next time step, $\hat{x}$. The two networks have the same number of layers and are connected diagonally at each layer such that the information encoded in the hidden units of the discriminative network is propagated to the corresponding layer of the generative network at the next time step. Intuitively, the diagonal connections allow higher levels of the network to represent high level phenomena, such as goals or individuality, while lower levels represent low level information, such as motion. Our experiments confirm this intuition. The model can be trained without any action labels, in which case the hidden state may be used to discover high level information about the data, or with action labels for a subset of the data, in which case each action is assigned to a hidden state unit and will thus contribute to subsequent motion prediction and action classification.

The flow of information through the network and the cost associated with its classification and prediction is expressed with the following functions: \\ 
\vspace{-3mm}
\begin{align*}
 & \text{ ~Discriminative}		 & 	& \text{Generative} 									& &\text{ ~~~~~~Cost}\\
h^1_i &= f([x_i, v_i],h^1_{i-1})    &	\hat{h}^L_{i+1} &= f(h^L_{i}, \hat{h}^L_{i}) 				& C_y &= \sum\nolimits_{i=1}^T \mathcal{L}_y(y_i,\hat{y}_i)\\
h^l_i  &= f(h^{l-1}_i,h^l_{i-1})     &  	\hat{h}^l_{i+1}  &= f(\hat{[h}^{l+1}_{i+1}, h^l_i], \hat{h}^l_i)		& C_x &= \sum\nolimits_{i=1}^T \mathcal{L}_x(x_{i+1},\hat{x}_{i+1}) \\
\hat{y}_i &= (h^L_i(1:N)+1)/2   	 &  	\hat{x}_{i+1}     &= g(\hat{h}^1_{i+1}) 					& C &= \lambda C_y + (1-\lambda) C_x\\
\end{align*}

where $f$ is recurrent cell function, $g$ is a transformation, and $\mathcal{L}_y$ and $\mathcal{L}_x$, are loss functions (see Figure \ref{mainfigure}). The total cost, $C$, combines the misclassification cost, $C_y$, and misprediction cost, $C_x$, using $\lambda$ to tradeoff the two. $N$ is the number of labeled actions, $L$ the number of levels, $T$ the number of frames in $x$, $l$ is the layer index and $i$ the frame index.
Classification labels $\hat{y}_i$ are assigned to the first $N$ units of $h^L_i$ and scaled from $[-1$ $1]$ to $[0$ $1]$. 

We present our model as part of a general framework where $f$, $g$, and number of levels/units are architectural choices to be optimized for each dataset. For our experiments we found that 2-3 levels of recurrent cells with 100-200 units worked well, with $f$ as a gated recurrent unit (GRU) cell \citep{cho2014learning} and $g$ as linear transformation. The choice of loss functions depends on the target type; sigmoid cross entropy for multitask classification (where actions can co-occur), softmax cross entropy for multiclass classification (where actions are mutually exclusive), and sum of squared differences for regression (where outputs are real valued). The optimal value for $\lambda$ depends both on the output domain of $\mathcal{L}_y$ and $\mathcal{L}_x$ and whether the primary goal is classification or simulation. Data-specific model- and training parameters are described in Section \ref{expanal} and further training details are discussed in \href{http://www.vision.caltech.edu/~eeyjolfs/behavior_modeling#trainingdetail}{supplementary material}\footnote{\url{www.vision.caltech.edu/~eeyjolfs/behavior_modeling}}.

%

\subsection{Multimodal prediction} 
\label{multimode}
Evidence suggests that animal behavior is nondeterministic \citep{roberts2016stochastic}, 
thus, motion prediction may be better represented as a probability distribution than a function. When future motion is multimodal, the best regression model will pick the average motion of the different modes which may not lie within any of the actual modes (visualized in \href{http://www.vision.caltech.edu/~eeyjolfs/behavior_modeling#multimodality}{supplementary material}). 
This observation has been made by others in the context of modeling real-valued sequences with RNNs, \citep{graves2013generating} model the output of an RNN as a Gaussian mixture model and \citep{chung2015recurrent} additionally model the hidden recurrent states as random variables.
We take a nonparametric approach, making no assumption about the shape of the distribution. We discretize motion into bins and treat the task of predicting future motion as independent multiclass classification problems for each motion feature, which results in a probability distribution over all bins for each dimension. 
More concretely, each dimension of $x$ is assigned $n$ bins and the target for $\hat{x}_{i+1}$ becomes the binned version of $x_{i+1}$, denoted as $\tilde{x}_{i+1}$, which has exactly one nonzero entry for each dimension of $x$. The prediction $\hat{x}_{i+1}$ then becomes a discrete distribution over the bins for each feature dimension and the motion prediction loss becomes $\mathcal{L}_x(x_{i+1}, \hat{x}_{i+1}) = \sum_d($crossentropy$(\tilde{x}_{i+1}, \hat{x}_{i+1}))$, as opposed to the Euclidean distance in the case when $x$ is a real valued vector.
The number of bins determines the granularity of the motor control; a greater number of bins means more precise motion control but is also more expensive to train. 

\subsection{Simulation} 
\label{simulation_section}
Given a model that can predict an agent's future motion from its current state, a virtual agent can be simulated by iteratively feeding predicted motion $\hat{x}_{i+1}$ as input $x_{i+1}$ to the network. 
We pick a bin by sampling from the distribution given by $\hat{x}_{i+1}$ and assign a real value to $x_{i+1}$ by sampling uniformly from the selected bin. 
An agent's perception of the environment depends on the agent's location, therefore, sensory features $v_{i+1}$ must be computed on the fly for each forward simulation step from the agents perspective at time $i+1$.
When simulating multiple agents that interact with one another, each agent is moved according to its $x_{i+1}$ and then $v_{i+1}$ is computed for each agent based on the new configuration of all agents.

\newpage
\section{Data} 
\label{datasection}
Our framework is \textit{agent centric}, it models the behavior of a single agent based on how it moves and experiences the world including other agents. It is applicable to any data that can be represented in terms of motor control (e.g. joystick controller), and sensory input that captures context from the environment (e.g. 1st person camera).
\
We test our model on two types of data, fruit fly behavior and online handwriting. Both can be thought of as a type of behavior represented in the form of trajectories, but the two are complementary: 
First, flies behave spontaneously, performing actions of interest sporadically and in response to its environment, while handwritten text is intensional and highly structured. Second, handwriting varies significantly between different writers in terms of size, speed, slant, and proportions, while inter-fly variation is relatively small. 
\
With the datasets selected for our experiments, listed below, we are interested in answering the following questions:
1) does motion prediction improve action classification, 2) can the model generate realistic simulations
(does it learn the sensory-motor control), 
and 3) can the model discover novel behavioral phenomena?

\textbf{Fly-vs-fly} \citep{eyjolfsdottir2014detecting} contains pairs of fruit flies engaging in 10 labeled courtship- and aggressive behaviors. We include this dataset in our experiments to see how our model compares with our previous action detection work which relies on handcrafted window features.

\textbf{FlyBowl} is a video of 10 male and 10 female fruit flies interacting and is labeled with male wing extensions which is part of their courtship behavior. With this dataset we were particularly interested in whether our model could simulate a virtual fly in a complex, dynamic environment.

\begin{figure}[h]
  \hspace{-1.5mm}
  \includegraphics[width=\textwidth]{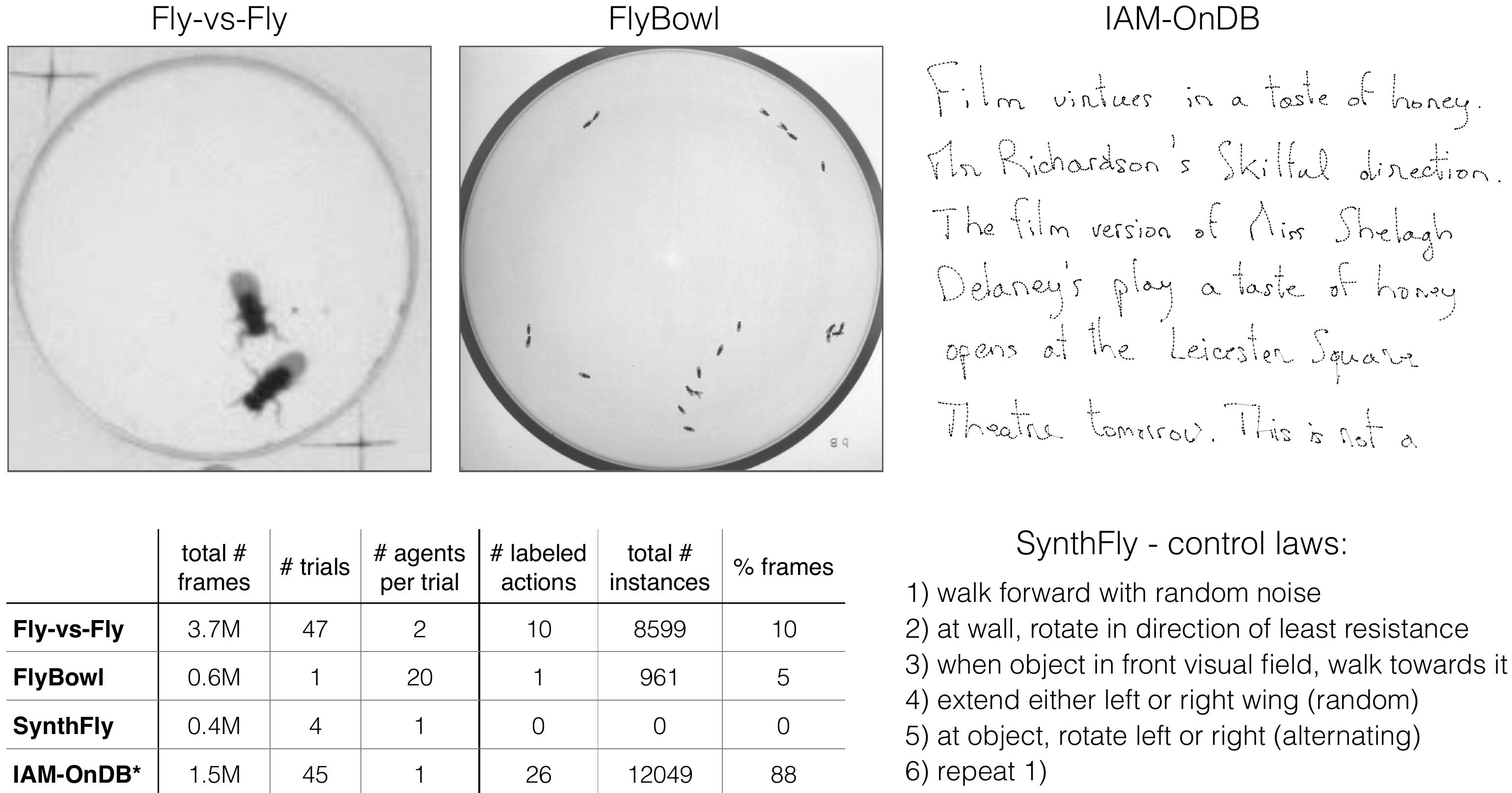}
  \caption{Snapshots from the three labeled datsets used for our evaluation and a list of control laws used to generate synthetic fly trajectories. The table summarizes the statistics of each experimental dataset, where \textbf{total \# frames} sums over all \textbf{trials} (videos / text documents) within an experiment and \textbf{agents} within a trial, \textbf{total \# instances} sums over all action classes, and \textbf{\% frames} is the percent of frames in labeled sequences containing actions of interest. IAM-OnDB* is a subset of IAM-OnDB with additional annotations for 10 of its trials.}
    \label{labeleddata}
\vspace{-1mm}
\end{figure}

\textbf{SynthFly} is a synthetic dataset containing a single fly moving inside of a rectangular chamber with a stationary object located in the center. The fly is synthesized to move according to the control laws listed in Figure \ref{labeleddata}. The purpose of this dataset is to test whether our model could learn generative control rules, particularly ones that enforce non-deterministic behavior (see laws 4 and 5). 

\textbf{IAM-OnDB} \citep{liwicki2005iam} contains handwritten text from 195 different writers, acquired using a smart whiteboard that records a list of (x, y) coordinates for each pen stroke. 
The data is weakly labeled, with each sequence separated into short lines of transcribed text. 
For consistency with our framework we hand annotated strokes of 10 writers, marking the start and end of the 26 lower case characters, which we use along with data from 35 unlabeled writers for our experiments. 

All data, along with details about training and test splits, will be available in \href{http://www.vision.caltech.edu/~eeyjolfs/behavior_modeling}{supplementary material}.

\begin{figure}
\vspace{-2mm}
  \includegraphics[width=\textwidth]{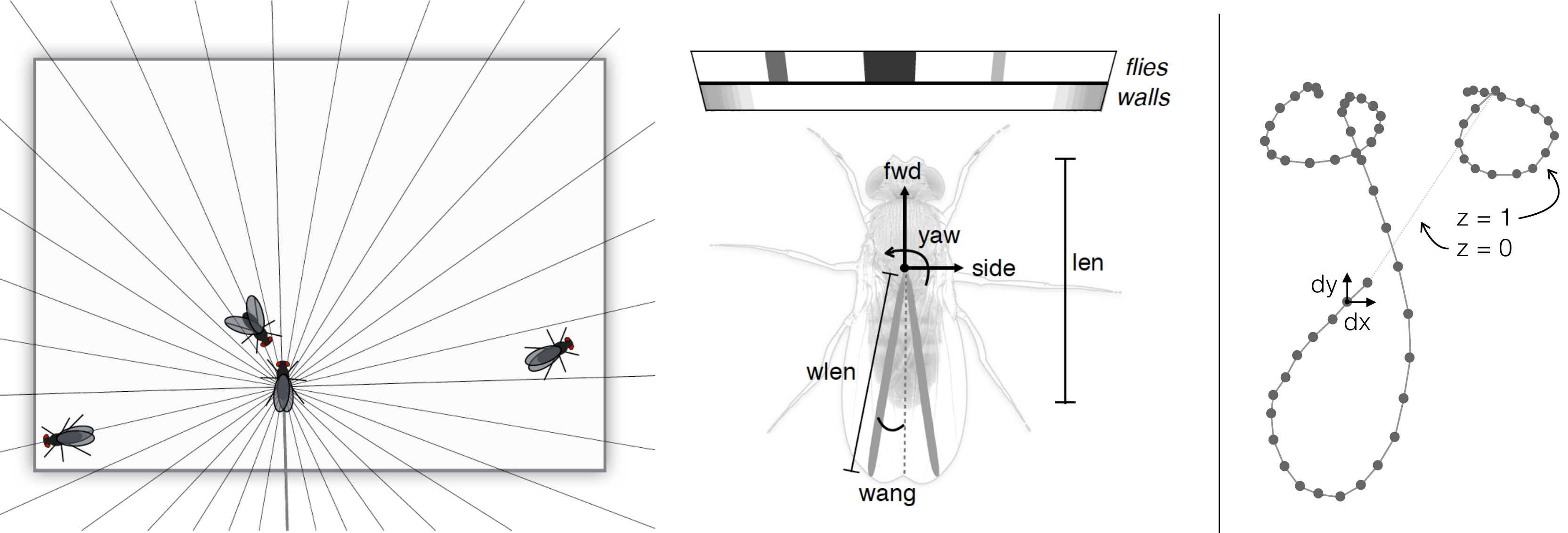}
  \caption{\textbf{Left:} Sensory input $v$ for fruit flies represents how a fly sees other flies and chamber walls, their motor control $x$ lets them move their body along 8 dimensions. \textbf{Right:} Motor control $x$ for handwriting is represented as vector (dx, dy) along with binary stroke visibility z (pen on/off whiteboard). }
    \label{sensorymotor}
\vspace{-2mm}
\end{figure}

\textbf{Fly representation:}
Motor control features, $x$, describe the locomotion of a fly. The flies are tracked from video using FlyTracker\footnote{\url{www.vision.caltech.edu/Tools/FlyTracker}} and from the tracked fly poses we extract motion features represented in the fly's frame of reference. The 8 motion features, displayed on top of the fly\footnote{Original photograph from \url{gompel.org/drosophilidae}} in Figure \ref{sensorymotor}, 
are designed such that they can animate virtual fly agents.
\
Sensory input features, $v$, are inspired by a fly's compound eye which consist of 750 compactly aligned ommatidia. Approximating its vision as a one dimensional 360$^{\circ}$ view, we place 72 5$^{\circ}$ circular sectors around a fly, aligned with its orientation, and project flies that overlap with a sector onto its artificial retina. Flies close to the agent yield high intensity in several ommatidia, and flies that are far away take up few ommatidia with low intensity. We represent chamber walls similarly, projecting them onto a separate channel decreasing intensity exponentially with distance to the fly. This representation is invariant of the shape of the chamber and the number of flies present in the chamber. 

In order to compare our model with methods presented in \cite{eyjolfsdottir2014detecting}, independently of feature representation, we use the 36 features provided with the Fly-vs-Fly dataset. We assign the first 8 dimensions which describe a fly's locomotion as motor control and the remaining features which describe its position relative to the other fly and feature derivatives as sensory input. 

\textbf{Handwriting representation:}
We represent the motor control, $x$, as (dx, dy, z) where dx and dy are the x and y displacements from the previous pen recording and z is a binary variable denoting segment visibility. We normalize dx and dy for each writer, providing invariance to writing speed, but character size (number of points per character), slant, and other variations are not explicitly accounted for. As handwriting is not influenced by a changing environment, but rather a function of the internal state and current motion of the writer, we leave the sensory input, $v$, empty. 

\section{Experiments and analysis}
\label{expanal}
We evaluate our framework on three objectives: classification,  simulation, and discovery. For classification we show the benefit of motion prediction as an auxiliary task, compare our performance on Fly-vs-Fly with previous work, and analyze the performance on IAM-OnDB. We qualitatively show that simulation results for fly behavior and handwriting look convincing, and that the model is able to learn control laws used to generate the SynthFly dataset. For discovery we show that hidden states of the model, trained only to predict motion (without any action labels), cleanly capture high level phenomena that affect behavior, such as fly gender and writer identity. 

\textbf{Model details:} We trained a separate model for each dataset, using a sequence length of 50, a batch size of 20, and 51 bins per dimension for motion prediction. For fly behavior data we used 2 levels of GRU cells (4 cells total) of 100 units each, and for handwriting we used 3 levels of GRU cells (6 cells total) of 200 units each. Parameters were determined using a rough parameter sweep on a subset of the training data. Further training details are described in \href{http://www.vision.caltech.edu/~eeyjolfs/behavior_modeling#trainingdetail}{supplementary material}. Our model is implemented in Tensorflow \citep{tensorflow2015-whitepaper}.

\subsection{Classification} 
From the sequence of frame-wise classifications we extract consecutive classifications into intervals (or \textit{bouts}) of actions. 
To measure both duration and counting accuracy we use the performance measures described in \cite{eyjolfsdottir2014detecting}, namely the F1 score (harmonic mean of precision and recall), on a per-frame and per-bout level. Bout-wise precision and recall is computed by assigning predicted bouts to ground truth bouts one-to-one, maximizing intersection over overlap. F* is the harmonic mean of the F1-frame and F1-bout scores.

\begin{figure}
 \vspace{-2mm}
  \includegraphics[width=390pt]{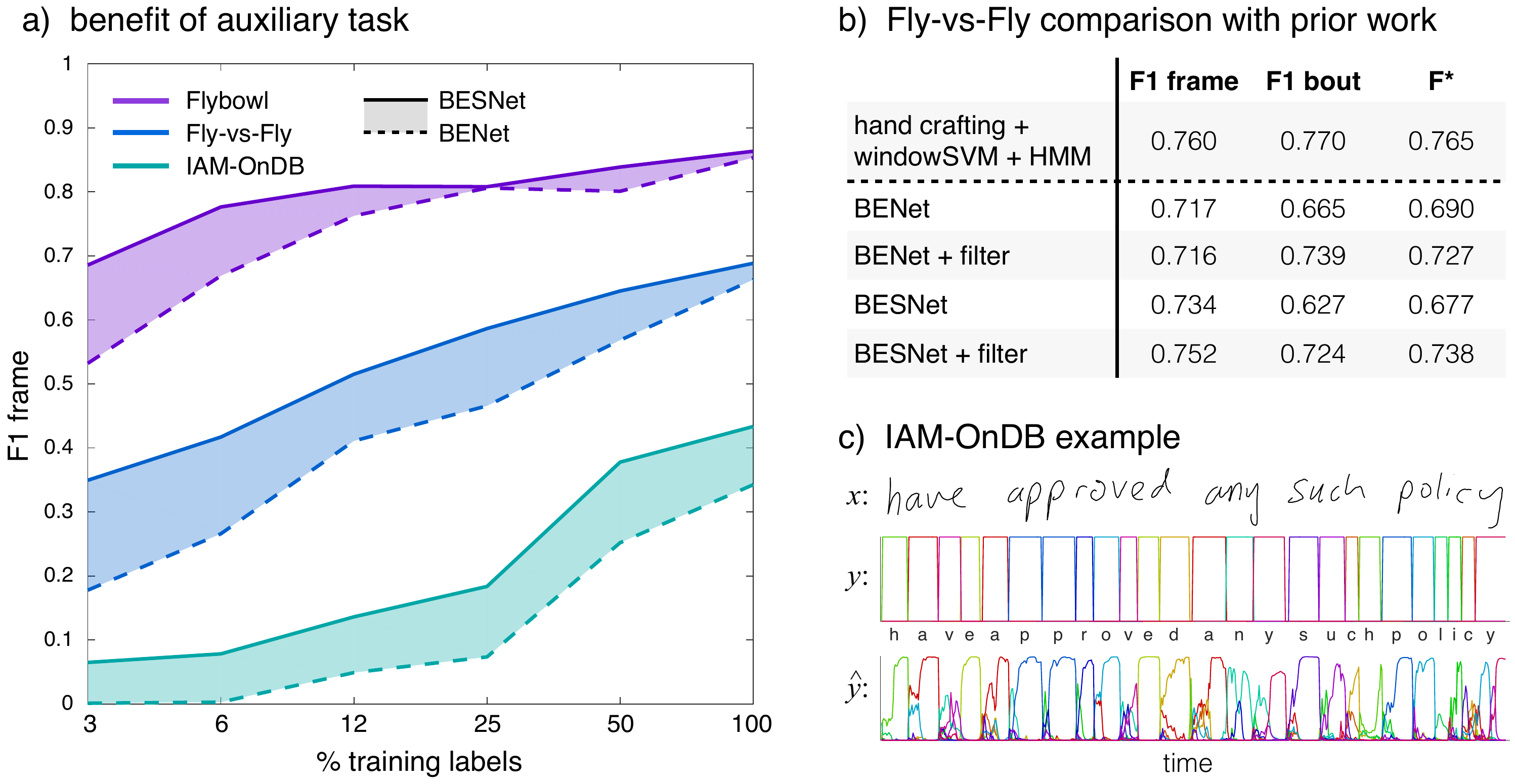}
  \caption{\textbf{a)} Performance of model trained with (solid, BESNet) and without (dashed, BENet) motion prediction, with varying number of training labels. \textbf{b)} Performance on Fly-vs-Fly compared with results from \cite{eyjolfsdottir2014detecting}. \textbf{c)} Example input $x$, ground truth $y$, and classification score $\hat{y}$ from IAM-OnDB.}
\label{classifig}
 \vspace{-2mm}
\end{figure}

Our main goal in terms of classification is to reduce the number of training labels without loss in performance, by using motion prediction as an auxiliary task. To measure the benefit of semi-supervision we compare our model, which we will refer to as Behavior Embedding Sensory-motor Network (BESNet), with our model without its generative part (similar to a standard RNN but with action labels embedded in hidden states, shown in Figure \ref{variants_loglik}), which we will refer to as Behavior Embedding Network (BENet). We trained both models on each dataset using 3-100\% of available labels. As BESNet is trained to predict future motion it makes use of unlabeled sequences during training whereas BENet does not. Figure \ref{classifig} a) shows the frame-wise F1 score for each of the 36 trained models (3 datasets, 6 label fractions, 2 model types), averaged over all action classes in a dataset. This experiment shows that motion prediction as an auxiliary task significantly improves classification performance, especially when labels are scarce.

In Figure \ref{classifig} b) we compare the performance of our network with the best performing method on \textbf{Fly-vs-Fly}, a window based support vector machine (SVM) that uses hand crafted window features and fits an HMM to the output for smoother classification -- outperforming sophisticated methods such as structured SVM. For this comparison we used the features published with the dataset as described in Section \ref{datasection}. 
Although recurrent networks implicitly enable smooth classification, different actions require different levels of smoothness. 
To avoid over segmentation of action intervals, we smooth the output of our network by applying a flat filter, of size equal to 10\% of the mean duration of each class. Our results show that filtering significantly improves the bout-wise performance and that our performance on the Fly-vs-Fly test set is comparable with that of \cite{eyjolfsdottir2014detecting}, using no handcrafting and no context of future frames (apart from smoothing). 

We applied the same type of filtering to the classification output of  \textbf{IAM-OnDB} as we did for Fly-vs-Fly and obtained an F1-(frame, bout) of (0.445, 0.585) averaged over all classes, and (0.567, 0.690) averaged over all instances (weighted average of classes).
Figure \ref{classifig} c) demonstrates that at the beginning of some characters there tends to be more confusion in $\hat{y}$ than towards the end, which is unsurprising as the beginning of these characters looks approximately the same. 

\subsection{Motion prediction} 
Before we look at simulation results, we quantitatively measure the accuracy of one-step predictions. We compute the log-likelihood of FlyBowl test sequences under the motion prediction model:  loglik($x) = \sum_{i=1}^{T-1} \sum_{d=1}^8$ log$(\tilde{x}_{i+1}^d \cdot {\hat{x}_{i+1}^d})$, where $\tilde{x}_{i+1}^d$ is the ground truth indicator vector for bins of motion dimension $d$, and $\hat{x}_{i+1}^d$ is a probability distribution over the bins predicted by the model. 

\begin{figure}[h]
\centering
\vspace{-1mm}
 \includegraphics[width=\textwidth]{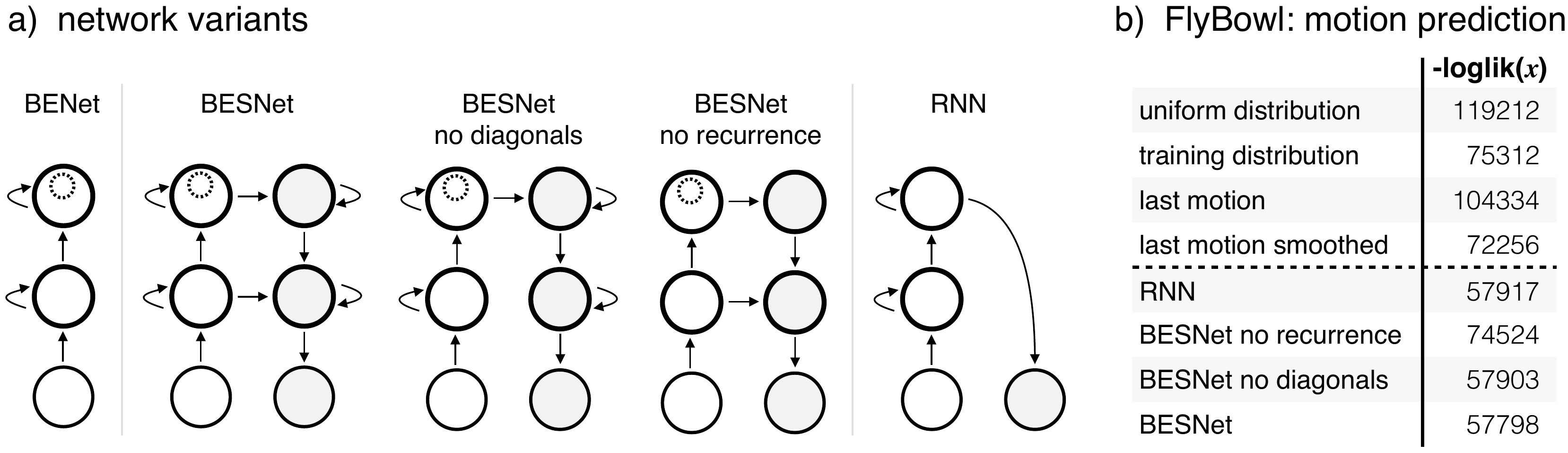}
\caption{\textbf{a)} Network variants used in experiments (compare with highlighted cells in Figure \ref{mainfigure}). 
\textbf{b)} Motion prediction performance on FlyBowl testset, see text for explanation.}
\label{variants_loglik}
\vspace{-2mm}
\end{figure}

We compare our model with the following motion prediction policies: 1) uniform distribution over bins, 2) distribution over bins computed from training set, 3) constant motion policy that copies previous indicator vector as motion prediction, and 4) a smooth version of 3) filtered using an optimized Gaussian kernel. The results, shown in Figure \ref{variants_loglik}, demonstrate that the recurrent models learn a significantly better policy. 
\
In addition, we compare variants of our model and a standard RNN within our framework (with the same sensory-motor representation, multimodal output, and GRU cells) which shows that recurrence is essential for good motion prediction and that diagonal connections provide a slight performance gain. 
In Section \ref{discovery} we show the main benefit of the diagonals.

\subsection{Simulation}
One-step prediction performance does not clearly reveal whether a model has learnt the generative process underlying the training data. In order to get a better notion of that we look at simulations produced by the learnt models, which can be thought of as very long term predictions. As motion prediction is probabilistic, comparing long term predictions with ground truth becomes difficult as the domain of probable positions becomes exponentially large. Qualitative inspection, however, gives a good intuition about whether the simulated agent has learnt reasonable control laws. 

\begin{figure}[h]
  \hspace{-1mm}
  \includegraphics[width=400pt]{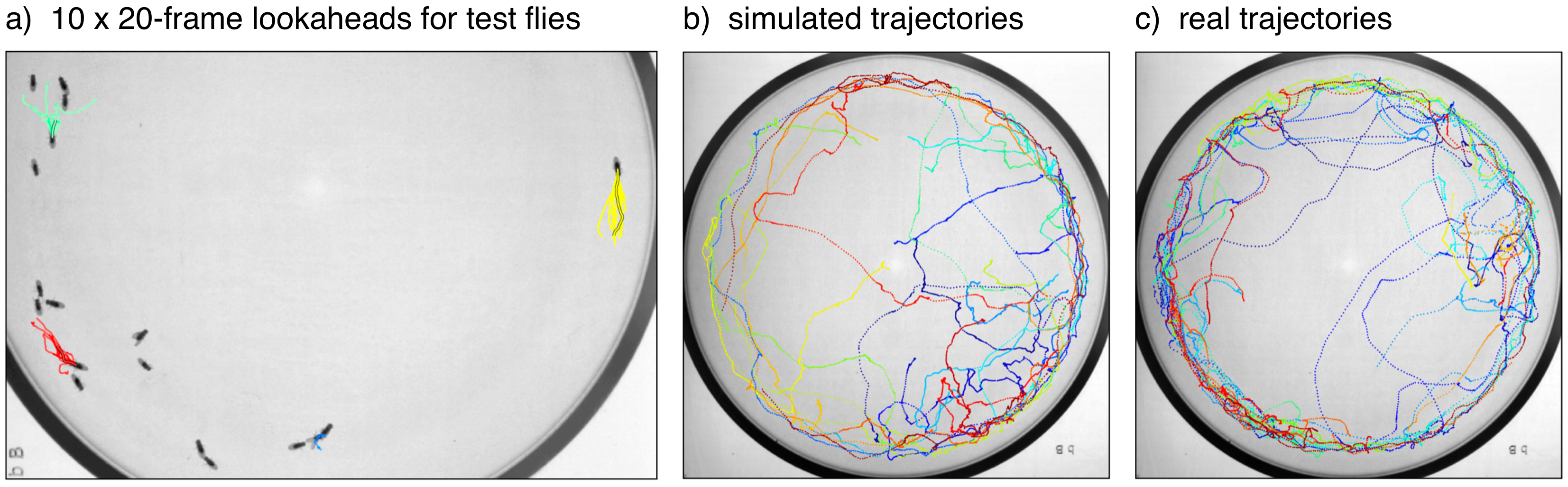}
  \vspace{-4mm}
  \caption{\textbf{a)} 10 x 20-frame lookaheads (simulations) for each test fly from its current location, demonstrating the non-deterministic nature of the motion prediction. The ground truth 20-frame future trajectory is outlined in black for comparison. \textbf{b)} shows trajectories of 20 flies simulated for 1000 frames, and \textbf{c)} shows 1000-frame trajectories for 20 real flies interacting. The simulation shows that the model has learnt a preference for staying near the boundary and to avoid walking through the boundary.}
\label{simflies}
  
  \vspace{5mm}
  \hspace{-2.5mm}
  \includegraphics[width=410pt]{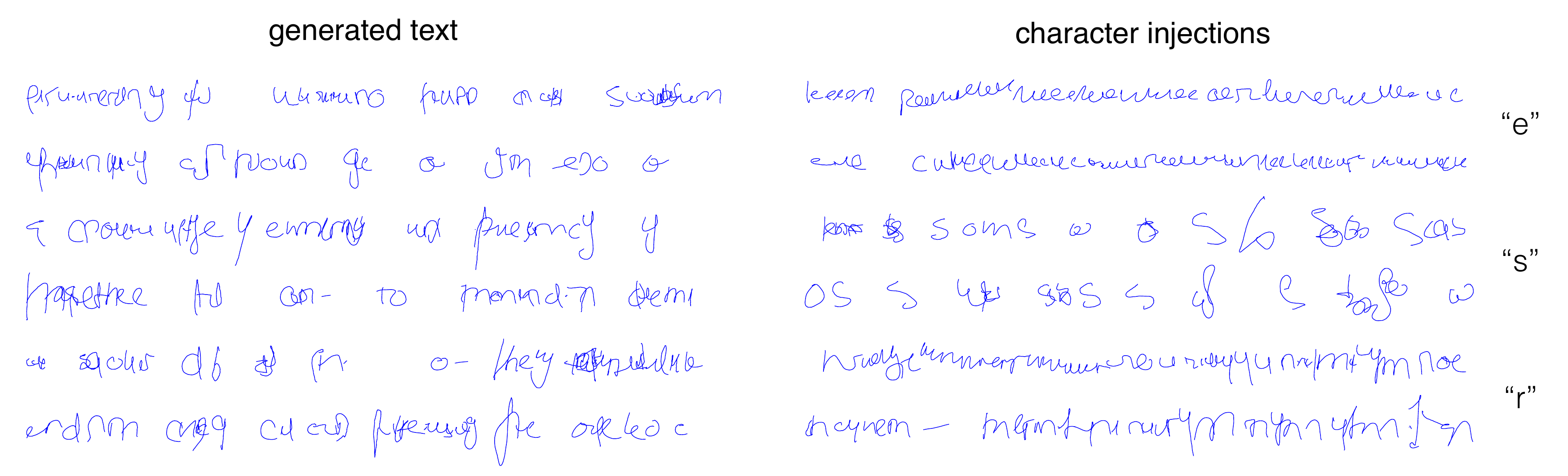}
  \vspace{-4mm}
  \caption[Handwriting simulation]{\textbf{Left:} Text generated by our model, one vector at a time (approximately 20 vectors per character). \textbf{Right:} Text generated by the same model while "activating" character classification units of the model during simulation, shown in two lines per character.}
    \label{textgen}
\end{figure}

While the underlying generative process for the motion of real flies is unknown, simulations from the model trained to imitate them suggest that the model has learnt a reasonable policy. During simulation we place no physical constraints on how the flies can move but our results show that simulated \textbf{FlyBowl} agents avoid collisions with the chamber walls and with other flies, and that agents are attracted to other flies and occasionally engage in courtship-like behavior. This is shown in Figure \ref{simflies} and better visualized as video in \href{http://www.vision.caltech.edu/~eeyjolfs/behavior_modeling#flybowlsim}{supplementary material}.

Simulated handwriting is easier to visualize in an image and we are used to recognizing the structure it should produce. Figure \ref{textgen} shows that the model trained on \textbf{IAM-OnDB} produces character-like trajectories in word-like combinations. Note that handwriting is generated one (dx, dy, z) vector at a time, and each character is composed of roughly 20 such points on average. 
On the right hand side of Figure \ref{textgen} we show that we can increase the generation of specific characters by activating their classification units (forcing their values to 1 and others to 0) during simulation.

\begin{figure}
  \includegraphics[width=400pt]{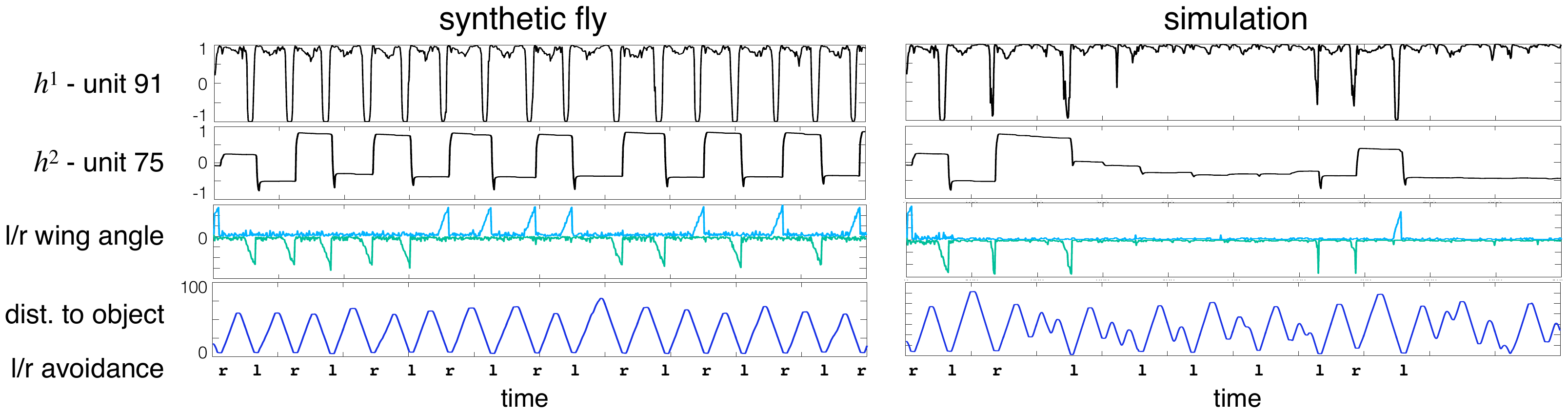} 
  \caption{Comparison between synthetic fly (ground truth) and simulation by our model. The wing angles, distance to object, and left/right turn show the agent's motion over time, and the two hidden units indicate that the model has learnt to represent control laws 4 and 5 used to generate the synthetic trajectories. }
    \label{synthsim}
\end{figure}

Figure \ref{synthsim} shows the output of two recurrent units of the \textbf{SynthFly} model that indicate that the model was able to learn control rules that were designed to ensure a multimodal motion prediction target. One unit fires in correlation with either left or right wing extension, and the other toggles between a negative and positive state as the agent turns left or right to avoid the object. In \href{http://www.vision.caltech.edu/~eeyjolfs/behavior_modeling#synthflysim}{supplementary material} we show a video of this simulation and compare it to a simulation from the model trained with deterministic motion prediction. This comparison clearly demonstrates the benefit of treating motion prediction as a distribution over bins, as the deterministic agent quickly becomes degenerate.

\subsection{Discovery} 
\label{discovery}
We motivated the structure of our network, specifically the diagonal connections between discriminative and generative cells, with the intuition that it would allow higher levels of the network to better represent high level phenomena. 
To verify this we train models to only predict future motion, with \textbf{no classification target}, and visualize what the hidden states capture. We apply the model to $[x, v]$, obtaining hidden state vectors $h^l$ and $\hat{h}^l$, $l \in \{1, ..., L\}$, and prediction $\hat{x}$, map the data points (time steps of each fly/writer) from each state to 2 dimensions using t-distributed stochastic neighbor embedding (tSNE, \cite{maaten2008visualizing}), and plot them in colors based on known phenomena. 

\begin{figure}
\includegraphics[width=400pt]{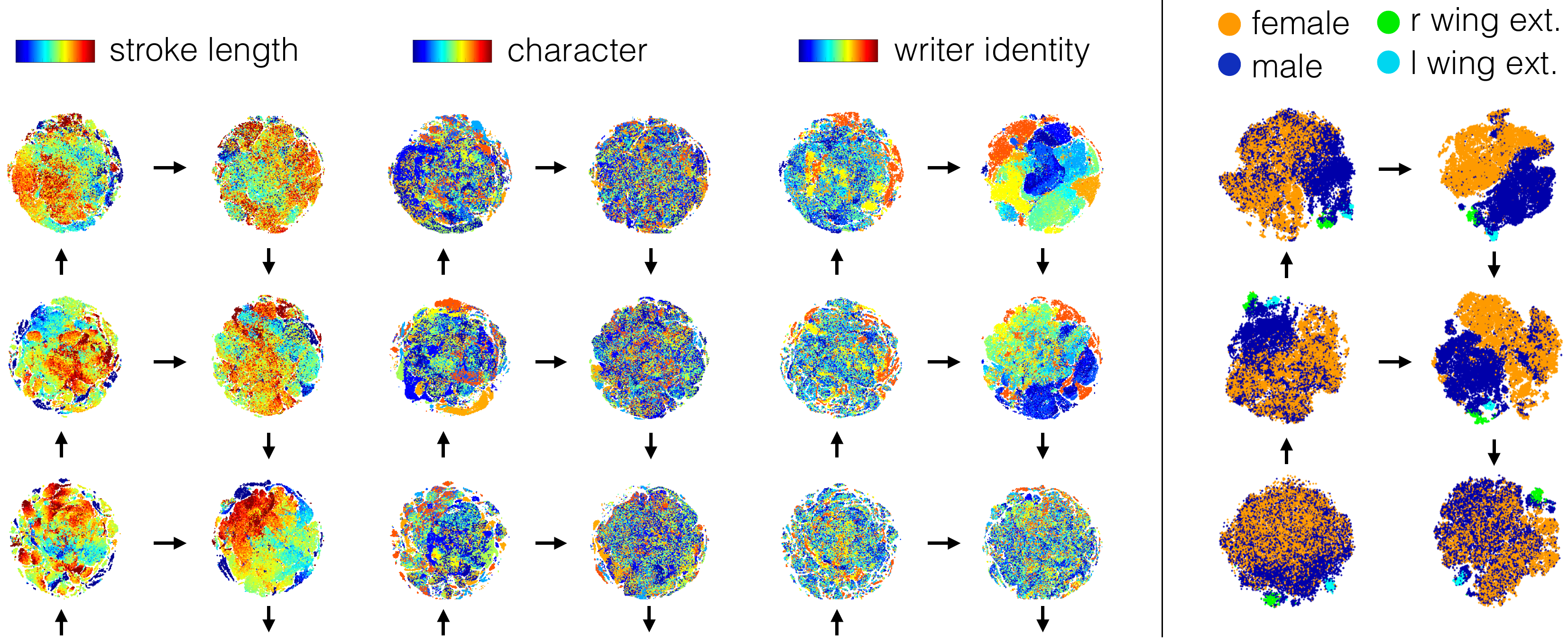}
  \caption{\textbf{Left:} Hidden state values of a 3 level model trained without any labels on IAM-OnDB,  reduced to 2 dimensions using tSNE mapping. The network discovers writer identity at the highest level, while lower level phenomena such as stroke length are represented at lower levels. 
\textbf{Right:} tSNE mapped input, output, and hidden state values of FlyBowl model (trained without any labels), colored by gender and male wing extension. 
}
\label{unsuper_cluster}
\end{figure}

In Figure \ref{unsuper_cluster} we plot the data points of a 3 level (L=3) model trained on \textbf{IAM-OnDB} in this low dimensional embedding, and color code them according to three criteria: stroke length, character class, and writer identity. The results show that stroke length is well clustered at low levels but not at high levels, characters are best clustered at mid to top discriminative levels, and writer identity is extremely well clustered at the top generative level but not at low levels. We ran the same experiment for the model trained without diagonal connections (which without a classification target is effectively a standard RNN with 6 levels of GRU cells), which did not learn to represent writer identity in any of its hidden states. Intuitively this is because the network has to carry low level information through every state to predict low level information at the other end, whereas BESNet carries it directly through the low level diagonal connections leaving higher hidden states free to capture high level information. A visualization comparing both models is shown in \href{http://www.vision.caltech.edu/~eeyjolfs/behavior_modeling#diagonal}{supplementary material} along with a quantitative measurement of our observation.

The same visualization for the model trained on \textbf{FlyBowl}, where data points are color coded by gender and left/right wing extension, shows (Figure \ref{unsuper_cluster}, right) that gender is very mixed in the input and output states but well separated in the top generative state, while lower level information such as wing extension is well represented at lower levels of the network.

\section{Conclusion} 
We have proposed a framework for modeling the behavior of animals, that simultaneously classifies their actions and predicts their motion. We showed empirically that motion prediction (a target that requires no labeling) is a good auxiliary task for training action classifiers, especially when labels are scarce. We also showed that the generative task can be used to simulate trajectories that look natural to the human eye, and that activating classification units increases the frequency of that action in the simulation. Finally, we showed that our model lends itself well to discovery of high level information from the data.

Moving forward, we are interested in working on hierarchical label embedding in the states, assigning higher order activities to units higher in the network. Along those lines, a discrete recurrent network could be trained separately on the wealth of available text, and be placed on top of a real-valued handwriting network. We also aim to explore how this framework can be used to understand the neural mechanisms underlying the generation of behavior in flies.

\newpage
\bibliography{references}

\begin{thebibliography}{26}
\providecommand{\natexlab}[1]{#1}
\providecommand{\url}[1]{\texttt{#1}}
\expandafter\ifx\csname urlstyle\endcsname\relax
  \providecommand{\doi}[1]{doi: #1}\else
  \providecommand{\doi}{doi: \begingroup \urlstyle{rm}\Url}\fi

\bibitem[Abadi et~al.(2015)Abadi, Agarwal, Barham, Brevdo, Chen, Citro,
  Corrado, Davis, Dean, Devin, Ghemawat, Goodfellow, Harp, Irving, Isard, Jia,
  Jozefowicz, Kaiser, Kudlur, Levenberg, Man\'{e}, Monga, Moore, Murray, Olah,
  Schuster, Shlens, Steiner, Sutskever, Talwar, Tucker, Vanhoucke, Vasudevan,
  Vi\'{e}gas, Vinyals, Warden, Wattenberg, Wicke, Yu, and
  Zheng]{tensorflow2015-whitepaper}
Mart\'{\i}n Abadi, Ashish Agarwal, Paul Barham, Eugene Brevdo, Zhifeng Chen,
  Craig Citro, Greg~S. Corrado, Andy Davis, Jeffrey Dean, Matthieu Devin,
  Sanjay Ghemawat, Ian Goodfellow, Andrew Harp, Geoffrey Irving, Michael Isard,
  Yangqing Jia, Rafal Jozefowicz, Lukasz Kaiser, Manjunath Kudlur, Josh
  Levenberg, Dan Man\'{e}, Rajat Monga, Sherry Moore, Derek Murray, Chris Olah,
  Mike Schuster, Jonathon Shlens, Benoit Steiner, Ilya Sutskever, Kunal Talwar,
  Paul Tucker, Vincent Vanhoucke, Vijay Vasudevan, Fernanda Vi\'{e}gas, Oriol
  Vinyals, Pete Warden, Martin Wattenberg, Martin Wicke, Yuan Yu, and Xiaoqiang
  Zheng.
\newblock {TensorFlow}: Large-scale machine learning on heterogeneous systems,
  2015.
\newblock URL \url{http://tensorflow.org/}.
\newblock Software available from tensorflow.org.

\bibitem[Anderson \& Perona(2014)Anderson and Perona]{anderson2014toward}
David~J Anderson and Pietro Perona.
\newblock Toward a science of computational ethology.
\newblock \emph{Neuron}, 84\penalty0 (1):\penalty0 18--31, 2014.

\bibitem[Baldi(2012)]{baldi2012autoencoders}
Pierre Baldi.
\newblock Autoencoders, unsupervised learning, and deep architectures.
\newblock \emph{ICML unsupervised and transfer learning}, 27\penalty0
  (37-50):\penalty0 1, 2012.

\bibitem[Berman et~al.(2014)Berman, Choi, Bialek, and
  Shaevitz]{berman2014mapping}
Gordon~J Berman, Daniel~M Choi, William Bialek, and Joshua~W Shaevitz.
\newblock Mapping the stereotyped behaviour of freely moving fruit flies.
\newblock \emph{Journal of The Royal Society Interface}, 11\penalty0
  (99):\penalty0 20140672, 2014.

\bibitem[Braitenberg(1984)]{braitenberg84}
Valentino Braitenberg.
\newblock \emph{Vehicles Experiments in Synthetic Psychology}.
\newblock MIT Press, 1984.

\bibitem[Cho et~al.(2014)Cho, Van~Merri{\"e}nboer, Gulcehre, Bahdanau,
  Bougares, Schwenk, and Bengio]{cho2014learning}
Kyunghyun Cho, Bart Van~Merri{\"e}nboer, Caglar Gulcehre, Dzmitry Bahdanau,
  Fethi Bougares, Holger Schwenk, and Yoshua Bengio.
\newblock Learning phrase representations using rnn encoder-decoder for
  statistical machine translation.
\newblock \emph{arXiv preprint arXiv:1406.1078}, 2014.

\bibitem[Chung et~al.(2015)Chung, Kastner, Dinh, Goel, Courville, and
  Bengio]{chung2015recurrent}
Junyoung Chung, Kyle Kastner, Laurent Dinh, Kratarth Goel, Aaron~C Courville,
  and Yoshua Bengio.
\newblock A recurrent latent variable model for sequential data.
\newblock In \emph{Advances in neural information processing systems}, pp.\
  2962--2970, 2015.

\bibitem[Eyjolfsdottir et~al.(2014)Eyjolfsdottir, Branson, Burgos-Artizzu,
  Hoopfer, Schor, Anderson, and Perona]{eyjolfsdottir2014detecting}
Eyrun Eyjolfsdottir, Steve Branson, Xavier~P Burgos-Artizzu, Eric~D Hoopfer,
  Jonathan Schor, David~J Anderson, and Pietro Perona.
\newblock Detecting social actions of fruit flies.
\newblock In \emph{Computer Vision--ECCV 2014}, pp.\  772--787. Springer, 2014.

\bibitem[Graves et~al.(2013)Graves, Mohamed, and Hinton]{graves2013speech}
Alan Graves, Abdel-rahman Mohamed, and Geoffrey Hinton.
\newblock Speech recognition with deep recurrent neural networks.
\newblock In \emph{Acoustics, Speech and Signal Processing (ICASSP), 2013 IEEE
  International Conference on}, pp.\  6645--6649. IEEE, 2013.

\bibitem[Graves(2013)]{graves2013generating}
Alex Graves.
\newblock Generating sequences with recurrent neural networks.
\newblock \emph{arXiv preprint arXiv:1308.0850}, 2013.

\bibitem[Hochreiter \& Schmidhuber(1997)Hochreiter and
  Schmidhuber]{hochreiter1997long}
Sepp Hochreiter and J{\"u}rgen Schmidhuber.
\newblock Long short-term memory.
\newblock \emph{Neural computation}, 9\penalty0 (8):\penalty0 1735--1780, 1997.

\bibitem[Kabra et~al.(2013)Kabra, Robie, Rivera-Alba, Branson, and
  Branson]{kabra2013jaaba}
Mayank Kabra, Alice~A Robie, Marta Rivera-Alba, Steven Branson, and Kristin
  Branson.
\newblock Jaaba: interactive machine learning for automatic annotation of
  animal behavior.
\newblock \emph{nature methods}, 10\penalty0 (1):\penalty0 64--67, 2013.

\bibitem[Liwicki \& Bunke(2005)Liwicki and Bunke]{liwicki2005iam}
Marcus Liwicki and Horst Bunke.
\newblock Iam-ondb-an on-line english sentence database acquired from
  handwritten text on a whiteboard.
\newblock In \emph{Document Analysis and Recognition, 2005. Proceedings. Eighth
  International Conference on}, pp.\  956--961. IEEE, 2005.

\bibitem[Maaten \& Hinton(2008)Maaten and Hinton]{maaten2008visualizing}
Laurens van~der Maaten and Geoffrey Hinton.
\newblock Visualizing data using t-sne.
\newblock \emph{Journal of Machine Learning Research}, 9\penalty0
  (Nov):\penalty0 2579--2605, 2008.

\bibitem[Mnih et~al.(2015)Mnih, Kavukcuoglu, Silver, Rusu, Veness, Bellemare,
  Graves, Riedmiller, Fidjeland, Ostrovski, et~al.]{mnih2015human}
Volodymyr Mnih, Koray Kavukcuoglu, David Silver, Andrei~A Rusu, Joel Veness,
  Marc~G Bellemare, Alex Graves, Martin Riedmiller, Andreas~K Fidjeland, Georg
  Ostrovski, et~al.
\newblock Human-level control through deep reinforcement learning.
\newblock \emph{Nature}, 518\penalty0 (7540):\penalty0 529--533, 2015.

\bibitem[Moore(2002)]{moore2002some}
J~Moore.
\newblock Some thoughts on the relation between behavior analysis and
  behavioral neuroscience.
\newblock \emph{The Psychological Record}, 52\penalty0 (3):\penalty0 261, 2002.

\bibitem[Murphy(2002)]{murphy2002dynamic}
Kevin~Patrick Murphy.
\newblock \emph{Dynamic bayesian networks: representation, inference and
  learning}.
\newblock PhD thesis, University of California, Berkeley, 2002.

\bibitem[Rasmus et~al.(2015)Rasmus, Berglund, Honkala, Valpola, and
  Raiko]{rasmus2015semi}
Antti Rasmus, Mathias Berglund, Mikko Honkala, Harri Valpola, and Tapani Raiko.
\newblock Semi-supervised learning with ladder networks.
\newblock In \emph{Advances in Neural Information Processing Systems}, pp.\
  3532--3540, 2015.

\bibitem[Roberts et~al.(2016)Roberts, Augustine, Lawton, Lindsay, Thiele,
  Izquierdo, Faumont, Lindsay, Britton, Pokala, et~al.]{roberts2016stochastic}
William~M Roberts, Steven~B Augustine, Kristy~J Lawton, Theodore~H Lindsay,
  Tod~R Thiele, Eduardo~J Izquierdo, Serge Faumont, Rebecca~A Lindsay,
  Matthew~Cale Britton, Navin Pokala, et~al.
\newblock A stochastic neuronal model predicts random search behaviors at
  multiple spatial scales in c. elegans.
\newblock \emph{eLife}, 5:\penalty0 e12572, 2016.

\bibitem[Rumenlhart et~al.(1986)Rumenlhart, Hinton, and
  Williams]{rumenlhart1986learning}
DE~Rumenlhart, Geoffrey~E Hinton, and Ronald~J Williams.
\newblock Learning internal representation by error propagation, parallel
  distributed processing.
\newblock \emph{Explor. Microstruct. Cognition}, 1:\penalty0 318--362, 1986.

\bibitem[Silver et~al.(2016)Silver, Huang, Maddison, Guez, Sifre, Van
  Den~Driessche, Schrittwieser, Antonoglou, Panneershelvam, Lanctot,
  et~al.]{silver2016mastering}
David Silver, Aja Huang, Chris~J Maddison, Arthur Guez, Laurent Sifre, George
  Van Den~Driessche, Julian Schrittwieser, Ioannis Antonoglou, Veda
  Panneershelvam, Marc Lanctot, et~al.
\newblock Mastering the game of go with deep neural networks and tree search.
\newblock \emph{Nature}, 529\penalty0 (7587):\penalty0 484--489, 2016.

\bibitem[Simon(1996)]{simon1996sciences}
Herbert~A Simon.
\newblock \emph{The sciences of the artificial}.
\newblock MIT press, 1996.

\bibitem[Siwicki \& Kravitz(2009)Siwicki and Kravitz]{siwicki2009fruitless}
Kathleen~K Siwicki and Edward~A Kravitz.
\newblock Fruitless, doublesex and the genetics of social behavior in
  drosophila melanogaster.
\newblock \emph{Current opinion in neurobiology}, 19\penalty0 (2):\penalty0
  200--206, 2009.

\bibitem[Tinbergen(1963)]{tinbergen1963aims}
Niko Tinbergen.
\newblock On aims and methods of ethology.
\newblock \emph{Zeitschrift f{\"u}r Tierpsychologie}, 20\penalty0 (4):\penalty0
  410--433, 1963.

\bibitem[Valpola(2015)]{valpola2015neural}
Harri Valpola.
\newblock From neural pca to deep unsupervised learning.
\newblock \emph{Adv. in Independent Component Analysis and Learning Machines},
  pp.\  143--171, 2015.

\bibitem[Wiltschko et~al.(2015)Wiltschko, Johnson, Iurilli, Peterson, Katon,
  Pashkovski, Abraira, Adams, and Datta]{wiltschko2015mapping}
Alexander~B Wiltschko, Matthew~J Johnson, Giuliano Iurilli, Ralph~E Peterson,
  Jesse~M Katon, Stan~L Pashkovski, Victoria~E Abraira, Ryan~P Adams, and
  Sandeep~Robert Datta.
\newblock Mapping sub-second structure in mouse behavior.
\newblock \emph{Neuron}, 88\penalty0 (6):\penalty0 1121--1135, 2015.

\end{thebibliography}
\bibliographystyle{iclr2017_conference}

\end{document}